\documentclass[letterpaper]{article} 
\usepackage[preprint]{aaai2027}  
\usepackage[hyphens]{url}  
\usepackage{graphicx} 
\usepackage{natbib}  
\usepackage{caption} 
\usepackage{algorithm}
\usepackage[noend]{algpseudocode}

\usepackage{newfloat}
\usepackage{listings}
\DeclareCaptionStyle{ruled}{labelfont=normalfont,labelsep=colon,strut=off} 
\floatstyle{ruled}
\newfloat{listing}{tb}{lst}{}
\floatname{listing}{Listing}

\usepackage{amsmath} 
\usepackage{amssymb}  
\usepackage{amsmath}
\usepackage{xcolor}
\usepackage{colortbl}
\usepackage{amssymb}
\usepackage{enumitem}
\usepackage{booktabs}
\usepackage{multirow}
\usepackage{makecell}
\usepackage{subcaption}
\usepackage{cleveref}

\title{Search-Aided Joint Agent-Environment Reinforcement Learning \\ for Robust Lifelong Multi-Agent Path Finding with Rotations}
\author {
    He Jiang\textsuperscript{\rm 1},
    Jingtian Yan\textsuperscript{\rm 1},
    Yulun Zhang\textsuperscript{\rm 1},
    Yimin Tang\textsuperscript{\rm 2},
    Tanishq Duhan\textsuperscript{\rm 3},\\
    Rishi Veerapaneni\textsuperscript{\rm 1},
    Guillaume Sartoretti\textsuperscript{\rm 3},
    Jiaoyang Li\textsuperscript{\rm 1}
}
\affiliations {
    \textsuperscript{\rm 1}Carnegie Mellon University\\
    \textsuperscript{\rm 2}University of Southern California\\
    \textsuperscript{\rm 3}National University of Singapore\\
    \{hej2,jingtiay,yulunz\}@andrew.cmu.edu, 
    yimintan@usc.edu,
    e1280621@u.nus.edu,\\ 
    vrishi@cmu.edu,
    guillaume.sartoretti@nus.edu.sg,
    jiaoyangli@cmu.edu
}

\begin{document}

\maketitle

\begin{abstract}
Lifelong Multi-Agent Path Finding (LMAPF) requires repeatedly planning collision-free paths for agents that continuously receive new goals upon reaching their current ones. While many learning-based planners have been proposed for LMAPF, most rely on oversimplified kinematic assumptions that may overlook motion constraints critical to real-world performance. In this work, we study a more realistic yet scalable LMAPF model derived from many real-world automated warehouse systems, termed LMAPF-R2, which incorporates robust safety constraints and in-place rotation constraints. These constraints substantially increase coordination difficulty for learning-based planners, particularly in highly constrained spaces. To address these challenges, we propose Search-Aided Joint Reinforcement Learning (SJRL). We first augment neural policies with Causal PIBT, a single-step search-based planner that resolves agents' collisions and propagates their intentions. We then introduce a unified RL formulation that jointly optimizes agent and environment policies, where the environment policy learns graph edge costs to provide global movement guidance via backward Dijkstra search. Experiments demonstrate that SJRL achieves significant improvements over the strong search-based planner, Causal-PIBT, across multiple high-density maps. We further validate SJRL in a challenging mixed-reality warehouse environment with 8 physical robots and 248 virtual robots. 

\end{abstract}


\section{Introduction}
\label{introduction}

Multi-Agent Path Finding (MAPF)~\cite{SternSoCS19} studies the problem of planning collision-free paths for multiple agents from start vertices to goal vertices on a given graph. Lifelong MAPF (LMAPF) extends this setting by continuously assigning new goals to agents once they reach their current ones. The objective is to maximize system throughput, defined as the average number of goals reached by all agents per timestep.

LMAPF underpins many real-world multi-agent systems, including smart manufacturing facilities, automated scientific laboratories, and virtual gaming environments. Automated warehouses, as one of the key driving domains for LMAPF research, are now widely deployed worldwide by companies such as Amazon and Ocado. As demand for these systems continues to grow, increasingly complex and large-scale LMAPF deployments are expected in the near future. Accordingly, although many successful search-based planners have been developed over the years~\cite{li2021lifelong,okumura2022priority,chen2024traffic,jiang2024WPPL}, there is growing interest in learning-based approaches due to their potential for greater expressiveness and scalability.

\begin{figure}[tb]
    \centering
    \includegraphics[width=0.95\linewidth]{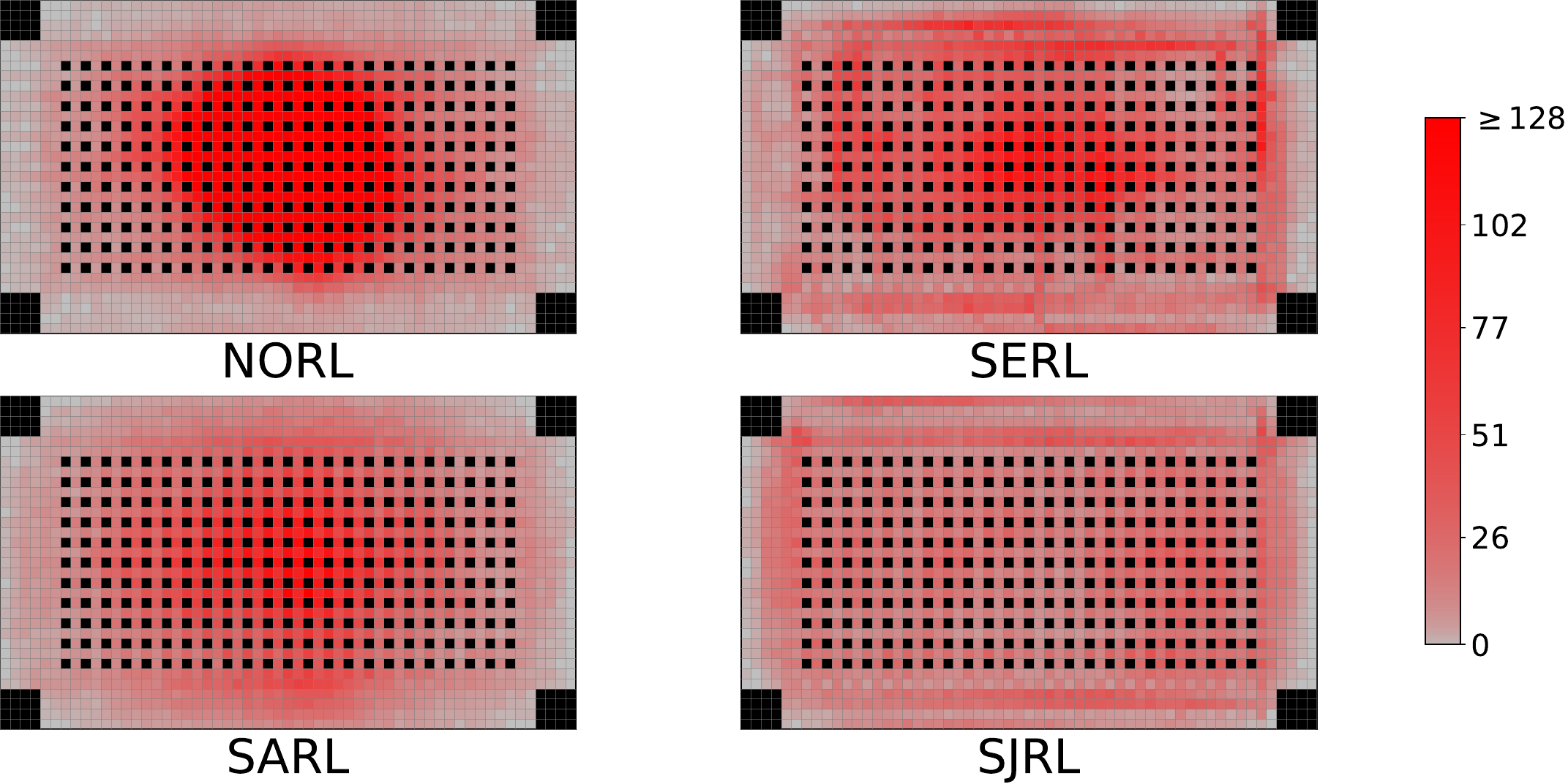}
    \caption{Heatmaps of average wait actions on a sortation map. Black cells denote obstacles; red intensity indicates wait frequency. NORL: no RL (Causal PIBT); SERL: environment RL; SARL: agent RL; SJRL: joint RL. SARL alleviates congestion locally, SERL balances traffic globally, and SJRL combines both. See~\Cref{ablation: joint learning} for details.}
    \label{fig:heatmap}
    \vspace{-0.1cm}
\end{figure}


Unlike many other multi-robot domains, LMAPF typically targets scenarios with \textbf{high agent densities}\footnote{Agent density is defined as the ratio of the number of agents to the number of free locations.}, \textbf{tightly constrained spaces, and long execution horizons}, as exemplified by automated warehouses. Consequently, both research and industrial LMAPF systems \textbf{often deliberately adopt grid graphs and relatively simple kinematic models to ensure stable and scalable coordination}. However, from the pioneering PRIMAL~\cite{sartoretti2019primal} and PRIMAL$_2$~\cite{damani2021primal} to more recent state-of-the-art methods such as MAPF-GPT~\cite{andreychuk2025mapf-gpt}, SILLM~\cite{jiang2025sillm} and HMAGAT~\cite{jain2026hmagat}, \textbf{nearly all learning-based LMAPF planners\footnote{Since planners for MAPF and LMAPF are largely adaptable, we do not explicitly distinguish between them unless necessary.} assume the oversimplified standard model~\cite{SternSoCS19}}, in which an agent moves directly to a neighboring vertex in a discrete timestep. Such oversimplification may fail to capture motion constraints that are critical for high-quality execution in real-world robotic systems.

In this work, we study a more realistic LMAPF model, motivated by many real-world automated warehouse systems, termed \textbf{LMAPF-R2}. It incorporates (1) \textbf{robust safety constraints} that enforce minimum safe distances between moving agents, and (2) \textbf{in-place rotation constraints} that capture the non-holonomic nature of many robots used in practice, such as differential-drive robots.  Both constraints have been previously studied for search-based planners. For example, prior work~\cite{atzmon2020robust} introduced robust MAPF and demonstrated its importance for reliable real-world execution, while studies such as~\cite{varambally2022mapf, zhang2023efficient, yan2025advancing} compared different modeling choices and highlighted the adverse consequences of neglecting rotation modeling during planning. 

Indeed, the modeling gap between planning and execution is equally critical for learning-based planners. However, to the best of our knowledge, it has received little attention in prior learning-based LMAPF studies. Thus, despite their remarkable progress under the standard model~\cite{andreychuk2025mapf-gpt2,andreychuk2025mapf-gpt,jiang2025sillm}, \textbf{we argue that learning-based LMAPF research should also embrace more realistic yet scalable models, such as LMAPF-R2, to keep pace with advances in search-based planning}.

In fact, robust and rotational constraints substantially increase the coordination difficulty for learning-based planners. The robust constraint requires agents to occupy additional vertices to ensure safe movement, while the rotational constraint forces agents to spend multiple timesteps to reach a neighboring vertex. As a result, the environment becomes more congested, and agents become more prone to deadlocks and livelocks. To address these challenges, we propose \textbf{Search-Aided Joint Reinforcement Learning (SJRL)}, built upon two key design principles: (1) leveraging the strengths of both search and learning, and (2) optimizing coordination from both agent and environment perspectives.

First, inspired by the success of Collision-Shield PIBT (CS-PIBT)~\cite{veerapaneni2024improving,jiang2025sillm}, which applies the single-step search algorithm PIBT~\cite{okumura2022priority} to resolve potentially colliding decisions of neural policies, we conjecture that such collision shielding can also benefit learning in the LMAPF-R2 setting. Accordingly, we integrate a variant of Causal PIBT~\cite{okumura2021time}, to accommodate the robust and rotational constraints and propagate agents' intentions.

Furthermore, motivated by the effectiveness of guidance graphs in coordinating agents through edge cost optimization in highly congested environments~\cite{jiang2024WPPL,zhang2024ggo,yukhnevich2025epibt}, we hypothesize that guidance graphs operate in a policy space complementary to the agent policy. Accordingly, we introduce a unified RL formulation that jointly optimizes the agent policy for local reactive coordination and the environment policy for global guidance generation.

Experiments under diverse settings show that joint optimization enables the two policies to reinforce each other, leading to consistent performance improvements. The heatmaps of wait actions on a sortation map in \Cref{fig:heatmap} exemplify how joint learning improves traffic flow.

Our main contributions can be summarized as follows:
\begin{enumerate}
[
    itemsep=1pt,
    parsep=2pt,
    topsep=2pt,
    partopsep=2pt
]
\item We are the first to study LMAPF-R2, a more realistic yet scalable kinematic model, for learning-based planners in scenarios with high agent densities and tightly constrained spaces, and advocate broader investigation into such models in learning-based planning.
\item To address the emerging coordination challenges, we propose a search-aided joint RL framework that learns both agent and environment policies, with tailored adaptations to collision shielding, guidance graph optimization, and network architecture.
\item Through extensive experiments, we demonstrate the importance of appropriate modeling, the benefits of joint learning, and the critical role of Causal PIBT in facilitating RL exploration by propagating agents' intentions.
\end{enumerate}

\section{Problem Formulation}

LMAPF-R2 studied in this work is a variant of LMAPF defined on a 4-neighbor grid graph $G = (V, E)$ with $n$ agents $A = \{a_1, a_2, \dots, a_n\}$. Vertices in $V$ represent traversable grid cells, and directed edges in $E$ connect adjacent vertices. Each agent is initialized at a unique start vertex $v \in V$ with an orientation $o \in \{\text{East}, \text{South}, \text{West}, \text{North}\}$.

Time is discretized into uniform timesteps. At each timestep, an agent may execute one of four actions: move forward to the adjacent vertex it faces, rotate $90^\circ$ clockwise, rotate $90^\circ$ counterclockwise, or wait. We prohibit both vertex collisions and following collisions. A vertex collision occurs when two agents occupy the same vertex at the same timestep. A following collision occurs when an agent moves into a vertex occupied by another agent in the previous timestep. Following collisions also include edge collisions, which occur when two agents swap their locations in a single timestep.

The first “R” in LMAPF-R2 denotes robust constraints, i.e., the prohibition of following collisions.\footnote{In $k$-robust MAPF~\cite{atzmon2020robust}, prohibiting following collisions is equivalent to 1-robust MAPF, which is sufficient for reactive neural policies.} Unlike standard LMAPF, where an agent may immediately follow another, LMAPF-R2 guarantees a minimum safe distance between agents. Thus, if the leading agent is delayed at the current timestep, the following agent can still safely execute its current action and respond to the delay in the next timestep. The second “R” denotes rotational constraints. Due to orientation, reaching an adjacent vertex may require one to three timesteps even in the absence of other agents.

LMAPF-R2 operates in a lifelong setting where agents are continuously assigned new goal vertices by an external task allocator whenever they reach their current goals. The system runs for a fixed time horizon, and the objective is to maximize throughput, defined as the average number of goals reached per timestep across all agents.

\section{Related Work}

Following the pioneering work PRIMAL~\cite{sartoretti2019primal}, numerous studies have explored learning-based planners for (L)MAPF. Some approaches learn policies from scratch using reinforcement learning (RL)~\cite{liu2020mapper, ma2021dcc, lin2023sacha}, others leverage imitation learning (IL) from expert search algorithms~\cite{li2021magat, andreychuk2025mapf-gpt, veerapaneni2025work, jiang2025sillm}, and some combine RL and IL~\cite{damani2021primal, wang2023scrimp}.

Despite extensive research on search-based planners that address robust constraints~\cite{atzmon2020robust,atzmon2020probabilistic,chen2021symmetry}, rotational constraints~\cite{zhang2023efficient,jiang2024WPPL,tao2025fast,yukhnevich2025epibt}, and more complex kinematic constraints~\cite{wen2022cl,yan2025mass,veerapaneni2025cbs-protocol} for real-world scenarios, few studies have investigated such constraints for learning-based (L)MAPF planners. One exception is~\cite{chan2022mapf-rot}, which extends PRIMAL to handle rotations with an ad hoc deadlock-avoidance mechanism based on action counting during inference. This work explores orthogonal methods that seamlessly integrate search algorithms into both training and inference, enabling efficient coordination under both robust and rotational constraints.

There has also been research on learning-based MAPF planners in continuous space for more complex kinodynamic settings, such as velocity and force control~\cite{dergachev2025corl,pshenitsyn2026camar,hu2025marf,huo2026mean}. However, as a tradeoff, these methods typically focus on relatively open environments with at most a few tens of agents and a one-shot task formulation. By contrast, we consider lifelong scenarios with high agent densities and long execution horizons.

Our agent policy learning builds upon prior research addressing collision resolution for neural policy outputs. Early methods froze agents’ movements upon detecting collisions~\cite{sartoretti2019primal}, which proved highly inefficient. More recently, CS-PIBT~\cite{veerapaneni2024improving} applied PIBT~\cite{okumura2022priority}, a lightweight search algorithm, to greedily resolve such collisions, and it has since become the de facto standard in state-of-the-art approaches~\cite{jiang2025sillm,jain2026hmagat,jain2026lagat}. Inspired by CS-PIBT, we integrate a variant of Causal PIBT~\cite{okumura2021time} into agent policy learning to efficiently handle robust and rotational constraints while resolving agents' collisions and propagating their intentions.

Our environment policy learning is adapted from guidance graph optimization (GGO)~\cite{zhang2024ggo}, which provides global guidance for agents’ movements. It formulates the guidance graph as edge costs on the map and applies CMA-ES~\cite{hansen2016cma}, a derivative-free evolutionary algorithm, to optimize either the edge costs directly or the parameters of a small neural network that generates them. In contrast, our approach formulates GGO within an RL framework and jointly trains it with the agent policy.

Joint optimization of agents and environments has rarely been explored for (L)MAPF. Prior work~\cite{gao2025co-optimization, li2025co-design} investigated the co-design of environment layouts and agent policies but was limited to small-scale scenarios involving at most 16 agents. In contrast, our work targets more practical settings by keeping the environment layout fixed and optimizing only the guidance graphs to coordinate hundreds of agents.

\section{Method}
\label{method}

We first present the Markov Decision Process (MDP) formulation in \Cref{method: mdp}, followed by agent policy learning with Causal PIBT in \Cref{method: agent policy} and environment policy learning with GGO in \Cref{method: environment policy}. Finally, we describe the joint RL procedure in \Cref{method: joint reinforcement learning}. Pseudocode and additional algorithmic details are provided in the appendix.

\begin{figure*}[!htbp]
    \centering
    \includegraphics[width=1.0\linewidth]{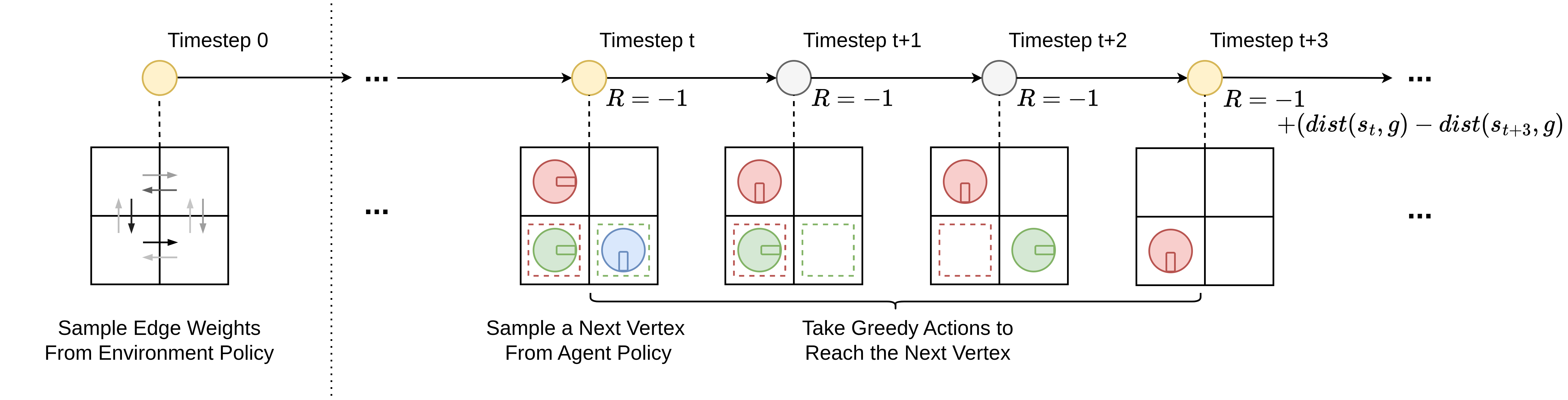}
    \vspace{-0.3cm}
    \caption{
    Markov Decision Process (MDP) from the perspective of the environment and the red agent. Upper circles represent states, and yellow ones denote states where a neural policy needs to make decisions. The circle in a grid represents a robot, with a small bar indicating its orientation. The dashed square of the same color as the robot indicates its subgoal. In this MDP, the environment first generates edge costs at timestep $0$, where darker arrows indicate lower costs and thus preferred movement directions. Agents begin to take actions at timestep $1$. Suppose the red agent is free at timestep $t$ and selects the vertex below it as its subgoal. It then becomes \textsc{busy} and moves greedily toward the subgoal unless its action is blocked by another agent (e.g., the green agent at timestep $t+1$, whose subgoal is the vertex to its right). Once the subgoal is reached, the agent becomes \textsc{free} again and needs to select its next subgoal (e.g., the red agent at timestep $t+3$). Additional details are described in~\Cref{method: mdp}.
    }
    \label{fig: mdp}
     \vspace{-0.1cm}
\end{figure*}

\begin{figure}[!btp]
    \centering
    \includegraphics[width=0.85\linewidth]{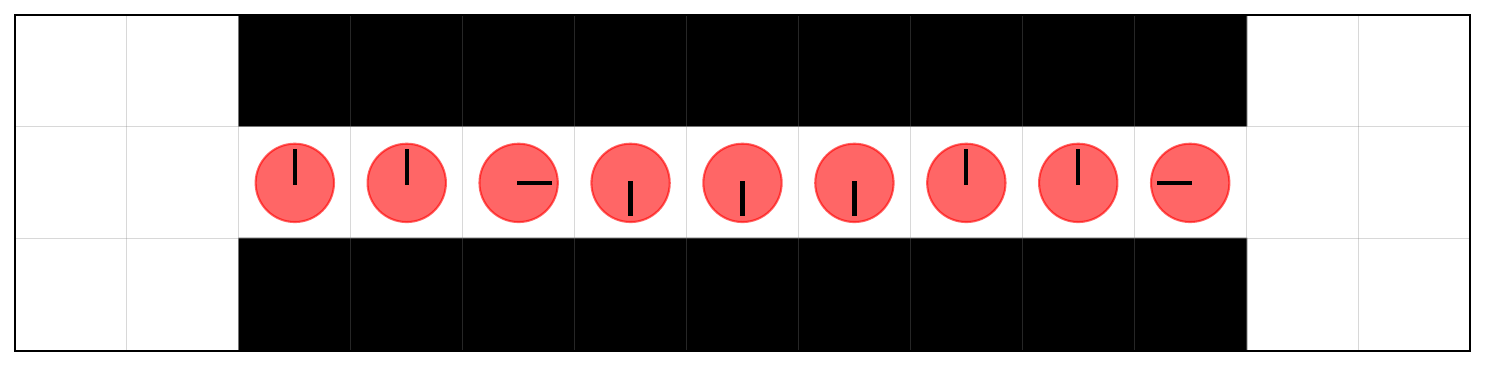}
    \vspace{-0.1cm}
    \caption{Illustration of the exploration challenge for nine agents in a long corridor under random exploration at the start of RL. Black and white cells denote obstacles and free space; red circles represent agents, and small black bars indicate their orientations.}
    \label{fig: long corridor}
    \vspace{-0.1cm}
\end{figure}

\subsection{MDP Formulation}
\label{method: mdp}
\Cref{fig: mdp} illustrates the MDP formulation of SJRL from the perspective of the environment and the red agent. For joint optimization, at timestep $0$, we first sample edge costs from the environment policy and then perform backward Dijkstra search to compute the minimum-cost distances between all pairs of states in the graph, where each state corresponds to a vertex–orientation pair. The resulting heuristic distances for states within an agent’s $11\times11$ local view are then provided as input to the agent policy network to guide movement, as detailed in \Cref{method: agent policy}.

Then, agents begin acting from timestep $1$. For an agent, if it is \textsc{free} at a timestep—namely, it has not yet decided its next vertex—we sample a next vertex from the agent policy as its subgoal. Causal-PIBT is then applied as post-processing to resolve collisions among subgoals and propagate agents' intentions, as detailed in \Cref{method: agent policy}. Once its subgoal is determined, the agent becomes \textsc{busy} and commits to reaching it using greedy actions. Due to the robust constraint, it may need to wait for other agents to move away before advancing to the next vertex. 

The agent receives a reward of $-1$ for each action taken to penalize timestep consumption. Upon reaching its subgoal, the agent becomes \textsc{free} again and selects a new subgoal. At this decision point, it also receives an additional reward of
$dist(s_{prev}, g) - dist(s_{curr}, g)$, where $s_{prev}$ and $s_{curr}$ denote the agent's state at the previous and current decision points, respectively. $dist(s, g)$ is the minimum number of timesteps required to reach the goal vertex $g$ from state $s$. It is worth noting that this distance is used for progress measurement, which is conceptually different from the previously computed heuristic distance used for movement guidance. This additional reward measures the progress made toward the goal since the previous decision. To encourage coordination, each agent additionally receives a team reward at every timestep equal to the average reward of all agents within its $5 \times 5$ local view. The reward for the environment policy is defined as the sum of the rewards of all agents.


\begin{figure*}[!htbp]
    \centering
    \includegraphics[width=0.9\linewidth]{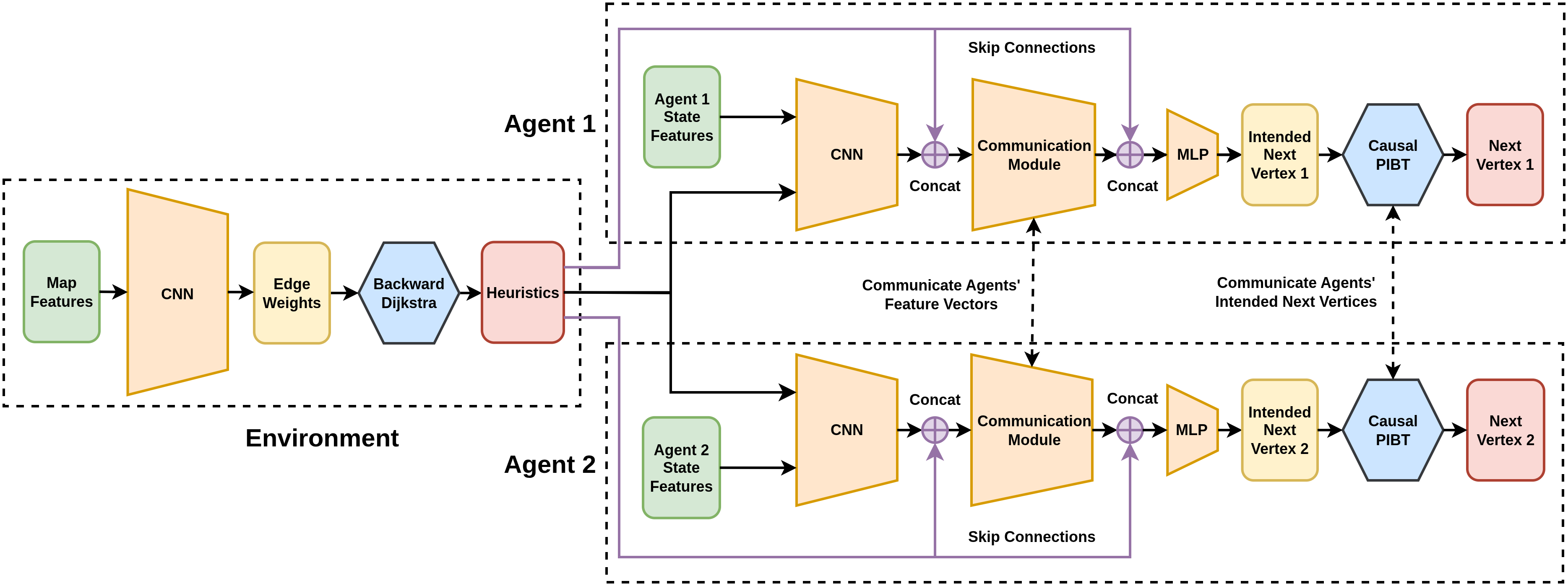}
    \caption{Neural policies for the environment and two communicating agents. Orange blocks denote learning modules, blue denotes search modules, green represents features, yellow represents decisions produced by neural policies, and red represents decisions post-processed by search. The purple lines indicate skip connections and the associated feature concatenations.}
    \label{fig: neural polices}
    \vspace{-0.1cm}
\end{figure*}

\subsection{Agent Policy Learning}
\label{method: agent policy}

For now, we assume that the environment policy is fixed and has already computed the heuristics used to guide movement. We now turn to the agent policy, illustrated in \Cref{fig: neural polices}. Our design builds upon the neural architecture of the state-of-the-art planner SILLM~\cite{jiang2025sillm}.

As in SILLM, the input to the neural network includes the locations of static obstacles within the agent’s local view, represented as a binary feature map. To facilitate collision resolution, we maintain a priority for each agent following PIBT~\cite{okumura2022priority}. These priorities are encoded as a ternary feature map, where $1$ indicates a higher-priority agent, $-1$ indicates a lower-priority agent, and $0$ denotes other locations. We also include an additional channel to indicate whether an agent is currently \textsc{busy} or \textsc{free}. Unlike SILLM, our setting considers four possible orientations at each location in the local view. Accordingly, we have a four-channel heuristic feature map, with each channel corresponding to one orientation.

In terms of architecture, we adopt the same convolutional neural network (CNN), communication module, and MLP as SILLM. In addition, we introduce two skip connections that directly inject heuristic information into the outputs of the CNN and the communication module. Specifically, these skip connections concatenate the learned feature vectors with the heuristic distances from the candidate next vertices to the goal. This design establishes a more direct dependency between the agent’s decisions and the heuristic guidance, thereby strengthening the influence of the environment policy and facilitating its optimization.

As described in the MDP formulation, the neural policy predicts the intended next vertex as a subgoal rather than a primitive action. We find that directly predicting primitive actions makes exploration substantially more challenging in multi-agent RL settings. Consider a sequence of agents in a long corridor (\Cref{fig: long corridor}): meaningful progress requires them to align in a common direction and move forward at certain timesteps. If all agents explore randomly during the early stages of RL training, the probability of achieving such coordination decreases exponentially with the number of agents in the corridor, rendering exploration prohibitively difficult. In contrast, predicting the next vertex as a subgoal and executing greedy actions toward it induces a simple policy hierarchy that enables higher-level coordination and more effective exploration.

We further replace the CS-PIBT~\cite{veerapaneni2024improving} used in SILLM with a synchronized variant of Causal PIBT~\cite{okumura2021time} to safeguard agents' intended next vertices. Unlike the original CS-PIBT, only agents that are \textsc{free} at the current timestep participate in joint decision-making, while the current and subgoal vertices of \textsc{busy} agents are treated as obstacles. In addition, we prohibit cyclic movements, as they would result in deadlocks during execution under the robust constraint. Aside from these modifications, our synchronized Causal PIBT follows the main logic of the CS-PIBT used in SILLM~\cite{jiang2025sillm}. Specifically, agents make decisions in priority order, employing a priority inheritance mechanism when a higher-priority agent attempts to move to a vertex currently occupied by a lower-priority agent~\cite{okumura19pibt,okumura2021time}. During decision-making, an agent first tries the highest-probability choice predicted by the neural policy, and then considers the remaining candidates in ascending order of their heuristic distances.

Notably, collision resolution for \textsc{free} agents considers only the robust constraints imposed by \textsc{busy} agents, but not those imposed by other \textsc{free} agents, since we allow a high-priority \textsc{free} agent to push away a low-priority \textsc{free} agent from its current vertex. Consequently, like other PIBT algorithms~\cite{okumura2022priority,okumura2021time,veerapaneni2024improving}, through depth-first search with priority inheritance, Causal PIBT can identify chains of \textsc{free} agents $a_1, a_2, \dots, a_k$, where agent $a_i$ moves to the current vertex of agent $a_{i+1}$ for all $i < k$, and $a_k$ moves to an unoccupied vertex. From another perspective, this process can be viewed as the propagation of $a_1$'s intention along the chain, as each $a_i$ pushes away $a_{i+1}$ from its current vertex. If such a chain lies in the long corridor shown in \Cref{fig: long corridor}, intention propagation can rapidly couple the decisions of the agents in the corridor, increasing the likelihood that they move in a common direction. As a result, the agents are more likely to form a coordinated movement pattern, enabling more efficient exploration of joint decisions.

After Causal PIBT is invoked, every agent is assigned a subgoal and becomes \textsc{busy}. Each agent then commits to reaching its subgoal by taking greedy actions: it first rotates until its orientation is correct, waits if the subgoal is occupied by another agent, and otherwise moves to the subgoal. If the subgoal coincides with its current vertex, the agent waits for one timestep before becoming \textsc{free} and making a new decision at the following timestep.

The benefits of predicting the next vertex as a subgoal to induce a hierarchical policy and applying Causal PIBT for intention propagation are demonstrated experimentally in \Cref{ablation: collision shielding}.

It is worth noting that Causal PIBT is applied during both training and inference. During training, it is treated as part of the environment, allowing PPO~\cite{schulman2017proximal} to optimize the policy directly without correcting policy gradients for the action modifications introduced by Causal PIBT.

\subsection{Environment Policy Learning}
\label{method: environment policy}

We now turn to the environment policy shown in \Cref{fig: neural polices}, which is implemented as a CNN consisting of five $3\times3$ convolutional layers with padding 1, ensuring that the final output has the same spatial dimensions as the input.

The input to the CNN includes the binary obstacle feature map and the simulation statistic feature maps proposed in~\cite{zhang2024ggo}. Specifically, we run an existing planner (in our case, the agent policy trained during the warm-up stage, described in \Cref{method: joint reinforcement learning}) to simulate LMAPF multiple times, record the visitation frequency of each vertex and the traversal frequency of each edge, and normalize these statistics as input features. Notably, these features are precomputed and remain fixed throughout both training and inference.

For each edge, the CNN outputs the mean and variance of a Gaussian distribution over its cost. Since each vertex has five outgoing edges (four directional moves and waiting), the CNN produces a total of $10$ output channels. Invalid edges are masked according to the graph structure. During training, edge costs are sampled from the predicted Gaussian distributions, whereas during inference, the means are used.

To bound edge costs within a predefined range, we apply an $\arctan$ transformation followed by a linear transformation, mapping each cost to the interval $[1,10]$. By adjusting the bias term of the prediction layer, we can initialize the mean edge costs to any value within this interval. For example, a bias of $0$ corresponds to an initial mean edge cost of $5.5$ after the transformation. The effect of manually initializing edge costs is analyzed in the appendix.

After the environment policy generates the edge costs, backward Dijkstra search is performed to compute the minimum-cost distances between all pairs of states in the graph. These distances are then provided as heuristic features to the agent policy, guiding the agents' decisions.

The environment policy can also be interpreted from another perspective: each edge is treated as a virtual agent that predicts its own cost. These virtual agents jointly shape the overall cost structure, which in turn determines the heuristic guidance for the movements of real agents. Accordingly, we adopt MAPPO~\cite{yu2022surprising} to train the environment policy, with all virtual agents sharing the same reward.

\subsection{Joint Reinforcement Learning}
\label{method: joint reinforcement learning}

Since the agent and environment policies are formulated within a unified RL framework, they can be trained jointly using shared simulations. However, random exploration by the agent policy during the early stages of training would introduce high variance into the optimization of the environment policy. To mitigate this issue, we first warm up the agent policy using uniform edge costs of $5.5$, and then proceed to joint training. Additional training details are provided in the appendix.

It is worth noting that, during inference, the edge costs and heuristic distances produced by the environment policy can be precomputed once for each map, requiring only the agent policy to be executed at each timestep. The edge costs can also be updated periodically to adapt to evolving traffic conditions, as in~\cite{zang2025online-ggo}; we leave this extension for future work.

\section{Experiments}

We conduct experiments on six maps with diverse obstacle structures (visualized in \Cref{fig:exp2_ablation_joint}) from the Moving-AI benchmark~\cite{SternSoCS19} and SILLM~\cite{jiang2025sillm}. For each map, we train a policy with $256$ agents and evaluate it with $32$--$320$ agents in increments of $32$. Each evaluation runs for $512$ timesteps. In all cases, the inference time per timestep is below $0.05$ seconds on an NVIDIA RTX 4090D GPU. Since maps are typically known in advance in real-world LMAPF applications, such as automated warehouses, we follow the state-of-the-art methods, SILLM~\cite{jiang2025sillm} and MAGAT+~\cite{jain2026lagat}, and focus on generalization across initial states, goal locations, and numbers of agents rather than across maps. Unless otherwise specified, we report results with $256$ agents, except for the main joint-learning experiment in \Cref{ablation: joint learning}. Additional results, including ablation on the network skip connections and validation with physical robots, are provided in the appendix.



\subsection{The Benefits of Joint Learning}
\label{ablation: joint learning}

\begin{figure}[!tb]
\centering
\input{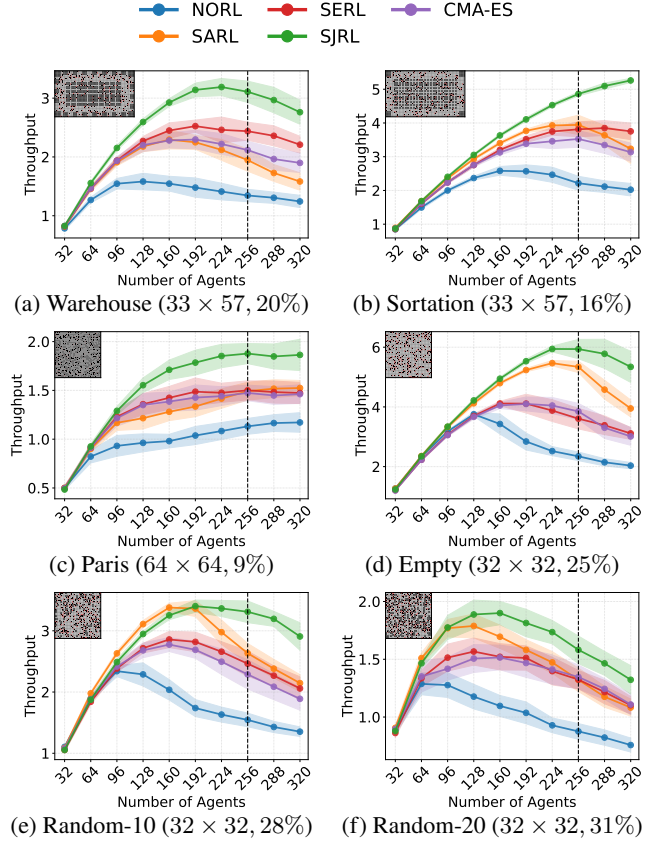}
\vspace{-0.1cm}
\caption{Ablation study on joint policy learning (\Cref{ablation: joint learning}). Curves show the mean throughput, with shaded regions indicating the standard deviation. Dashed vertical lines indicate the number of agents used for training. Maps with 256 randomly placed agents are shown in the top-left corner of each subfigure (best viewed when zoomed in). Black and gray cells denote obstacles and free space, respectively; red circles represent agents, and small black bars indicate their orientations. The size and the agent density of each map are shown in the parentheses following its name.}
\vspace{-0.1cm}
\label{fig:exp2_ablation_joint}
\end{figure}

We first compare our SJRL with the following:
\begin{enumerate}[
    itemsep=1pt,
    parsep=2pt,
    topsep=2pt,
    partopsep=2pt
]
    \item NORL: No RL.\\ Causal-PIBT~\cite{okumura2021time}, a strong search-based baseline that relies entirely on handcrafted heuristics for planning.
    
    \item SARL: Search-Aided Agent RL.\\ Only the agent policy is trained, with uniform edge costs as the guidance graph.
    
    \item SERL: Search-Aided Environment RL.\\ Only the environment policy is trained to generate the guidance graph, with Causal-PIBT as the agent planner.
    
    \item CMA-ES: the original GGO method~\cite{zhang2024ggo}.\\ The guidance graph is optimized using CMA-ES instead of RL, with Causal-PIBT as the agent planner.
\end{enumerate}

The results in \Cref{fig:exp2_ablation_joint} show that SJRL significantly outperforms the strong search-based baseline, Causal-PIBT, particularly as the number of agents increases. SJRL also consistently outperforms SARL and SERL, demonstrating the benefit of jointly optimizing the agent and environment policies and suggesting that these two policies contribute to LMAPF coordination in a complementary manner.

SERL achieves performance comparable to CMA-ES overall, while performing slightly better on the Warehouse, Sortation, and Random-10 maps. These findings demonstrate the potential of optimizing the guidance graph through reinforcement learning with more complex network structures.

\begin{figure}[!tb]
\centering
\input{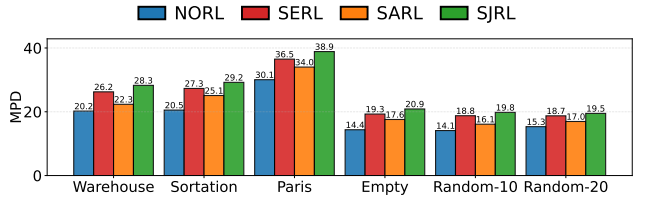}
\vspace{-0.4cm}
\caption{Comparison of mean pairwise distance (MPD) between different policies (\Cref{ablation: joint learning}).}
\vspace{-0.1cm}
\label{fig:utilization}
\end{figure}

We also compare different methods in \Cref{fig:utilization} using the mean pairwise distance (MPD), defined as
\begin{equation*}
\mathrm{MPD} = \sum_{i,j} p_i p_j \operatorname{dist}_{ij},
\end{equation*}
where $p_i$ is the empirical probability that vertex $i$ is occupied by some agent, and $\operatorname{dist}_{ij}$ is the shortest-path distance between vertices $i$ and $j$. Both the agent and environment policy learning increase the MPD, indicating that agents are more spatially dispersed and therefore experience less congestion.



\subsection{Comparison with State-of-the-Art Methods}
\label{model selection}

\begin{figure}[!tb]
\centering
\input{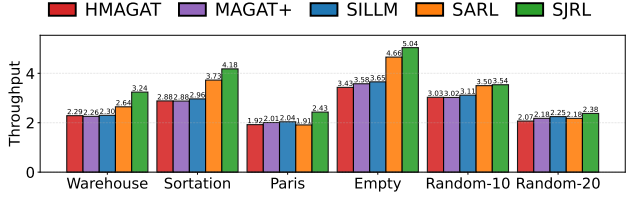}
\vspace{-0.4cm}
\caption{Comparison between SARL, SJRL and other state-of-the-art methods (\Cref{model selection}).}
\label{fig:adg_eval}
\vspace{-0.2cm}
\end{figure}

We further compare SARL and SJRL with other state-of-the-art (SoTA) methods, including SILLM~\cite{jiang2025sillm}, MAGAT+~\cite{jain2026lagat}, and HMAGAT~\cite{jain2026hmagat}, in \Cref{fig:adg_eval}, using differential-drive robots as in~\cite{yan2025advancing} and the Action Dependency Graph (ADG)~\cite{honig2019ADG} as the execution framework. The latter two SoTA methods are adapted to the LMAPF setting following the best practices in SILLM. SARL outperforms other SoTA methods on four of the six maps and performs on par with them on the remaining two, while SJRL consistently outperforms all of them. A key reason is that SARL and SJRL adopt the LMAPF-R2 model, whereas the other methods use the standard LMAPF model. The results suggest that, although ADG can accommodate diverse robot kinematics during execution, there is a need to study more realistic yet scalable LMAPF models, such as LMAPF-R2, for learning-based planners.

In principle, the other three SoTA methods, which rely on IL, could also be adapted to the LMAPF-R2 model. However, our preliminary investigation suggests that such an adaptation would require non-trivial modifications to multiple components, including the expert planner, collision-shielding mechanism, and training algorithm, and may lead to non-negligible performance degradation compared with the standard model. We therefore leave the adaptation of IL methods to the LMAPF-R2 model for future work.

\subsection{Policy Hierarchy and Intention Propagation}
\label{ablation: collision shielding}

\begin{figure}[!tbp]
\centering
\input{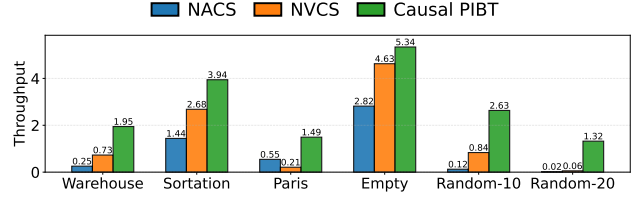}
\vspace{-0.4cm}
\caption{Comparison between methods with different decision spaces and collision-shielding strategies (\Cref{ablation: collision shielding}).}
\label{fig:exp1_throughput_comparison}
\vspace{-0.2cm}
\end{figure}

Finally, we compare our variant of Causal-PIBT with two alternative straightforward adaptations of CS-PIBT as collision-shielding mechanisms for agent RL:
\begin{enumerate}
[
    itemsep=1pt,
    parsep=2pt,
    topsep=2pt,
    partopsep=2pt
]
    \item Naive Action-Based Collision Shielding (NACS): the agent policy predicts primitive actions rather than next vertices, and a naive priority-based collision-shielding method is applied to resolve action conflicts.
    \item Naive Vertex-Based Collision Shielding (NVCS): the agent policy predicts the next vertices, and a naive priority-based collision-shielding method is applied to resolve vertex conflicts.
\end{enumerate}

The naive priority-based collision shielding in the robust setting works as follows. Regardless of whether the agent policy predicts actions or next vertices, all \textsc{free} agents make decisions sequentially in their priority order:
\begin{enumerate}
[
    itemsep=1pt,
    parsep=2pt,
    topsep=2pt,
    partopsep=2pt
]
    \item If a vertex was occupied in the previous timestep, no agent can take this vertex for the next movement in the current timestep due to the robust constraint.
    \item If a higher-priority agent has taken a vertex for the next movement, a lower-priority agent can no longer take this vertex for the next movement in the current timestep.
\end{enumerate}

Both NACS and NVCS ensure safe plans, but unlike Causal-PIBT, they cannot propagate agents' intentions through depth-first search and priority inheritance.

As shown in \Cref{fig:exp1_throughput_comparison}, NACS underperforms NVCS in most cases, likely due to the lack of a policy hierarchy. Moreover, both methods perform substantially worse than Causal-PIBT, with the largest performance gaps on the warehouse and random maps. We attribute this primarily to the exploration difficulty discussed in \Cref{method: agent policy}. Specifically, the warehouse map contains multiple aisle segments of length $3$, while the random maps contain long, narrow corridors, both of which make coordinated exploration particularly challenging. In contrast, our variant of Causal-PIBT benefits from the policy hierarchy and intention propagation, largely mitigating this difficulty. These results further suggest that straightforward adaptations to more complex robot kinematics may lead to poor performance, underscoring the importance of carefully designing and studying new LMAPF models for learning-based planners.



\section{Conclusion}

Achieving strong performance in real-world robotic systems requires balancing abstraction and realism in problem modeling. In this work, we study a more realistic yet scalable model for learning-based LMAPF planners, termed LMAPF-R2, which incorporates robust and rotational constraints. 

To address the coordination challenges introduced by LMAPF-R2, we propose SJRL, guided by two key principles: (1) leveraging the complementary strengths of search and learning, and (2) optimizing coordination from both the agent and environment perspectives. 

We hope this work encourages future research on increasingly realistic yet scalable problem formulations for learning-based LMAPF, facilitating the deployment of learning-based planners in real-world applications.





\clearpage

\bibliography{aaai2027}


\clearpage

\appendix

This appendix provides additional algorithm pseudocode, training details, experimental results, and physical-robot validation that were omitted from the main paper due to space limitations.

\section{Algorithm Pseudocode}

\begin{algorithm}[t]
\caption{Synchronized Causal PIBT}
\label{alg:SC-PIBT}
\begin{algorithmic}[1]
\State // For simplicity, the following required data are global variables that can be accessed by PIBT.
\Require Agent priority $p_i$, goal $g_i$, status $s_i$ 
\Require Agent vertex $v_i$, orientation $o_i$, subgoal $sg_i$
\Require Distance heuristics $h$
\Require Subgoal vertex $bv_i$ predicted by the agent policy
\Ensure Update $sg_i$ for \textsc{IDLE} agents
\Ensure Select actions for all agents.
\State // $O_b$: Vertices occupied by BUSY agents
\State // $O_c$: Vertices occupied currently by IDLE agents
\State // $O_n$: Vertices occupied next by IDLE agents
\State Initialize empty hash maps $O_b$, $O_c$, and $O_n$, mapping each occupied vertex to its corresponding agent index.
\For{each agent $i$}
    \If{$s_i=\textsc{IDLE}$}
        \State $O_c[v_i]=i$
        \State $sg_i=\textsc{NULL}$
    \Else
        \State $O_b[v_i]=i$
        \State $O_b[sg_i]=i$
    \EndIf
\EndFor
\For{each IDLE agent $i$ in priority order}
    \If{$sg_i==\textsc{NULL}$}
        \State \Call{PIBT}{$i$,$i$,$v_i$,$O_b$,$O_c$,$O_n$}
    \EndIf
\EndFor 
\For {each agent $i$}
    \State Greedily select an action for agent $i$ toward its subgoal $sg_i$, subject to the robust constraint.
        \label{line:action_selection}
\EndFor
\end{algorithmic}
\end{algorithm}

\begin{algorithm}[t]
\caption{PIBT Recursion}
\label{alg:PIBT}
\begin{algorithmic}[1]
\Require current agent index $i$
\Require root agent index $r$
\Require root agent vertex $rv$
\Require Vertices occupied by BUSY agents $O_b$
\Require Vertices occupied currently by IDLE agents $O_c$
\Require Vertices occupied next by IDLE agents $O_n$
\Ensure Update $sg_i$ for agent $i$ and return if PIBT succeeds
\Function{PIBT}{$i$,$r$,$rv$,$O_b$,$O_c$,$O_n$}
    \State Sort candidate next vertices for agent $i$, by prioritizing the vertex $bv_i$ predicted by the agent policy and then ranking the remaining vertices in ascending order of their heuristic distances to the goal according to $h$.
    \For{$nv$ in sorted candidate next vertices}
        \State // cyclic movement
        \If{$i==r$ and $nv==rv$} continue 
            \label{line:cyclic_movement}
        \EndIf
        \State // collision with a BUSY agent 
        \If{$nv\in O_b$} continue
            \label{line:collision_busy}
        \EndIf 
        \State // vertex collision with an IDLE agent
        \If{$nv\in O_n$} continue
        \EndIf
        \State // edge collision with an IDLE agent
        \If{$nv\in O_c$}
                \label{line:edge_collision_s}
            \State $j=O_c[nv]$
            \If{$j\neq i \land sg_j==v_i$} continue
                \label{line:edge_collision_e}
            \EndIf
        \EndIf
        \State // reserve the vertex $nv$ as the subgoal for agent $i$
        \State $O_n[nv]=i$
        \State $sg_i=nv$ 
            \label{line:update_subgoal1}
        \State // priority inheritance
        \If{$nv\in O_c$}
            \State $j=O_c[nv]$
            \If{$j\neq i \land sg_j==\textsc{NULL}$}
                \State $succ=$\Call{PIBT}{$j$,$r$,$rv$,$O_b$,$O_c$,$O_n$}
                \If{$\lnot succ$} continue
                \EndIf
            \EndIf
        \EndIf
        \State \Return True
    \EndFor
    \State // take the wait action as the fallback
    \State $O_n[v_i]=i$
    \State $sg_i=v_i$
        \label{line:update_subgoal2}
    \State \Return False
\EndFunction
\end{algorithmic}
\end{algorithm}

We describe the pseudocode of our Causal PIBT variant in ~\Cref{alg:SC-PIBT,alg:PIBT}. Since the algorithm closely follows the original PIBT~\cite{okumura19pibt} and the pseudocode includes detailed comments, we do not explain it line by line. Instead, we highlight two key points of the algorithm, which are also the main differences from the original PIBT.

\begin{enumerate}
\item The PIBT procedure in our algorithm plans subgoals for \textsc{IDLE} agents rather than their primitive actions (\Cref{alg:PIBT}, Line~\ref{line:update_subgoal1} and~\ref{line:update_subgoal2}). During planning, both the current and subgoal vertices of \textsc{BUSY} agents are treated as obstacles (\Cref{alg:PIBT}, Line~\ref{line:collision_busy}). After PIBT reasoning, each agent greedily selects its primitive action toward its assigned subgoal (\Cref{alg:SC-PIBT}, Line~\ref{line:action_selection}).
\item The PIBT procedure largely ignores the robust constraint to facilitate intention propagation among agents by considering only edge collisions rather than following collisions (\Cref{alg:PIBT}, Line~\ref{line:edge_collision_s}-\ref{line:edge_collision_e}). The only exception is that cyclic movements are disallowed, as allowing them would cause deadlocks during execution (\Cref{alg:PIBT}, Line~\ref{line:cyclic_movement}). The robust constraint is instead enforced primarily during the subsequent action selection stage (\Cref{alg:SC-PIBT}, Line~\ref{line:action_selection}).
\end{enumerate}

The computational complexity of synchronized Causal PIBT is identical to that of PIBT up to constant-factor overhead introduced by checking the robust constraints and busy-agent obstacles. 

\section{Training Details}
The warm-up training stage of the agent policy consists of 16 iterations, each with 64 simulations. The subsequent joint training stage consists of 128 iterations, each with 128 simulations. Due to GPU memory constraints, during warm-up, we update the agent policy every 32 timesteps, although each simulation lasts 512 timesteps. During joint training, the agent policy is updated after each simulation, but only 32 of the 512 timesteps are sampled for learning. The environment policy does not face this limitation because it makes decisions only once at the beginning of each simulation.

For the agent policy, the critic network uses the same architecture as the policy network in~\Cref{method: agent policy}. For the environment policy, no critic network is used, and we directly normalize returns with their means and standard deviations as advantages.

Our policies are trained using the PPO variants described in~\Cref{method}. \Cref{tab: hyperparameters} lists the key training hyperparameters, while additional details can be found in the experiment configurations provided in the source code. These hyperparameters were selected through a grid search around commonly used best-practice values, subject to computational constraints.

The policies are trained on servers with 64 vCPUs (Intel® Xeon® Platinum 8481C), 4 RTX 4090D GPUs (24~GB each), and 320~GB of memory. On average, for each map, the warm-up training stage takes approximately 3 hours, and the joint training stage takes approximately 6 hours. In all cases, the mean inference time per timestep is below 0.05 seconds.

\begin{table}[!tbh]
    \centering
    \caption{Key training hyperparameters.}
    \label{tab: hyperparameters}
    \begin{tabular}{lcc}
        \toprule
        \textbf{Hyperparameter} & \textbf{Agent} & \textbf{Environment} \\
        \midrule
        Learning rate           & $5\times10^{-5}$ & $5\times10^{-4}$ \\
        Entropy loss weight     & $0$              & $0$ \\
        Clipping ratio          & $0.2$            & $0.1$ \\
        GAE $\lambda$           & $0.95$           & -- \\
        Discount factor $\gamma$& $0.99$           & -- \\
        Batch size              & $512$ steps      & $64$ simulations \\
        Max gradient norm       & $1.0$            & $1.0$ \\
        Max KL divergence       & $0.02$           & $0.02$ \\
        \bottomrule
    \end{tabular}
\end{table}

\section{Experiments}
By default, all experiment settings in this paper are evaluated by $32$ runs with different random seeds, and we report the mean and standard deviation. Goal vertices for the agents are generated independently and uniformly at random from the set of all vertices in the graph.

\begin{figure}[!tb]
\centering
\input{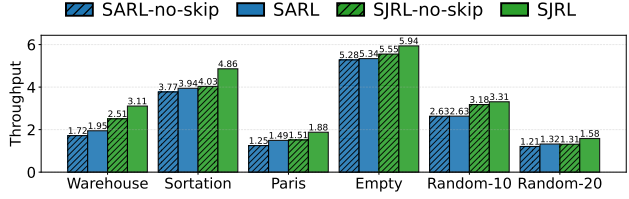}
\vspace{-0.6cm}
\caption{Throughput comparison between policies with and without skip connections (\Cref{ablation: skip connections}).}
\label{fig:exp3_throughput_comparison}
\end{figure}

\subsection {Ablation on Skip Connections}
\label{ablation: skip connections}
We compare the network architectures with and without skip connections in \Cref{fig:exp3_throughput_comparison}. Our results show that the skip connections improve the performance of both SARL and SJRL, but are more critical for SJRL. These results support our design choice of introducing skip connections to establish stronger dependencies between agent decisions and heuristic guidance, thereby facilitating guidance graph optimization in joint reinforcement learning.

\subsection{The Effect of Guidance Graph Initialization}
\label{exp: weight initialization}
For the warehouse and sortation maps with one-cell-wide aisles, human experts typically assign handcrafted edge costs within the aisles to encourage one-way traffic. Specifically, the preferred directions alternate between consecutive rows (or columns) of aisles.

We investigate whether SJRL can benefit from such expert priors. For these experiments, we initialize the edge cost of moving in the preferred direction to $3$, the edge cost of moving in the opposite direction to $7$, and all remaining edge costs to $5.5$, as in the default setting.

\begin{figure}[!tbp]
\centering
\input{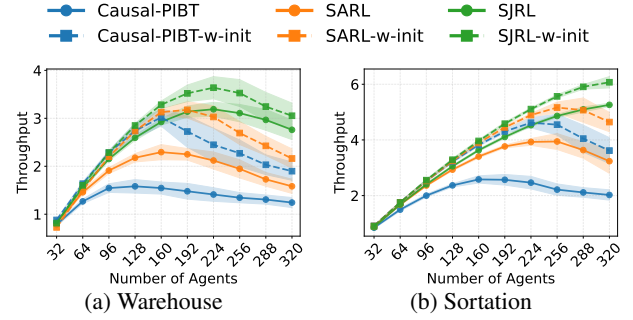}
\vspace{-0.3cm}
\caption{Throughput comparison between uniform edge-cost initialization (solid lines) and expert edge-cost initialization (dashed lines) on warehouse and sortation maps (\Cref{exp: weight initialization}). Curves show mean throughput, and shaded regions indicate the standard deviation.}
\label{fig:exp4_throughput_comparison}
\end{figure}

The results in~\Cref{fig:exp4_throughput_comparison} show that Causal-PIBT, SARL, and SJRL all benefit substantially from expert initialization, with weaker planners obtaining larger performance gains. However, SJRL trained from scratch still fails to match the performance of SJRL with expert initialization, suggesting that it may become trapped in a suboptimal local optimum during training, potentially due to the local optimization nature of PPO. This observation indicates that developing more effective optimization paradigms for joint learning is a promising direction for future work.

\subsection{Physical-Robot Validation}
Due to hardware and software limitations, we validate our algorithm using $8$ physical robots and $248$ virtual robots in a challenging mixed-reality warehouse environment with multiple aisles. Specifically, we use the OptiTrack Motion Capture System to localize the physical robots and the P3GASUS Framework~\cite{P3GASUS} to handle errors arising from execution disturbances and control inaccuracies. The throughput of different methods is compared in~\Cref{tab:mixed-reality}.

\textit{A video demo is provided in the supplementary material.}

\begin{table}
    \centering
    \caption{Comparison of throughput in a mixed-reality warehouse environment.}
    \label{tab:mixed-reality}
    \begin{tabular}{ccc}
    \toprule
    \textbf{Algorithm} & \textbf{Mean} & \textbf{Std} \\
    \midrule
    NORL  & 0.72 & 0.04 \\
    SERL  & 0.84 & 0.07 \\
    SARL  & 0.87 & 0.06 \\
    SJRL  & 1.01 & 0.07 \\
    \bottomrule
    \end{tabular}
\end{table}

\subsection{Comparison with State-of-the-Art Methods}
\label{model selection full}

\begin{figure}[!tb]
\centering
\input{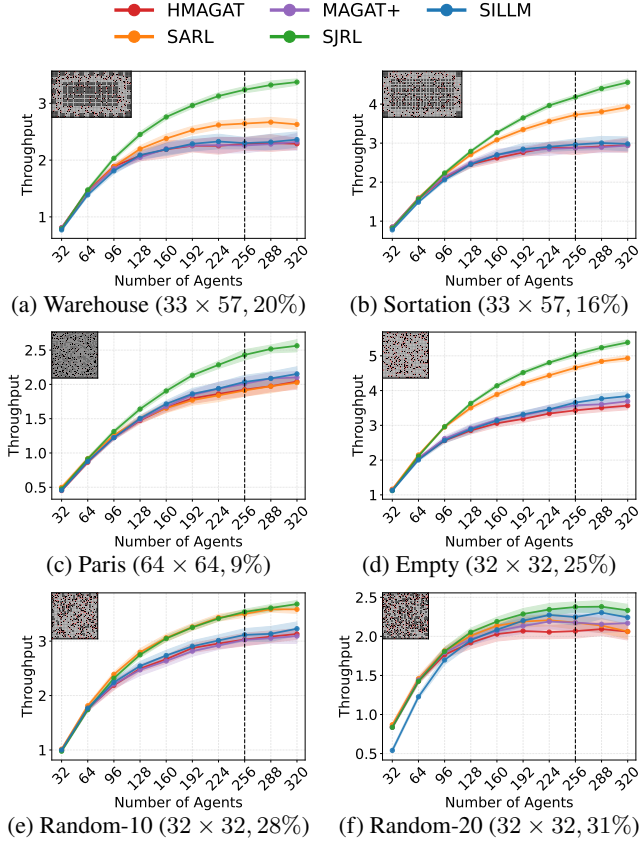}
\vspace{-0.1cm}
\caption{Comparison between SARL, SJRL and other state-of-the-art methods (\Cref{model selection full}). Curves show the mean throughput, with shaded regions indicating the standard deviation. Dashed vertical lines indicate the number of agents used for training. Maps with 256 randomly placed agents are shown in the top-left corner of each subfigure (best viewed when zoomed in). Black and gray cells denote obstacles and free space, respectively; red circles represent agents, and small black bars indicate their orientations. The size and the agent density of each map are shown in the parentheses following its name.}
\vspace{-0.1cm}
\label{fig:exp0_comparison_sota_full}
\end{figure}

In~\Cref{model selection}, we report only the mean throughput for the 256-agent setting. The complete results for different numbers of agents are provided in~\Cref{fig:exp0_comparison_sota_full}, and the conclusions are consistent with those reported in the main paper.

It is worth noting that, under the standard MAPF model, the state-of-the-art methods HMAGAT, MAGAT+, and SILLM achieve very similar performance in most cases. This suggests that recent advances in communication and representation learning have substantially narrowed the performance gap under the standard model. A possible explanation is that these methods share two key design principles: (1) communication mechanisms that effectively capture relative spatial relationships among agents, and (2) deeper CNN- or GNN-based architectures that stack multiple layers to provide strong representation capacity for processing the communicated information. 

Since this work directly adopts SILLM's communication module, an important direction for future work is to investigate more effective communication mechanisms for LMAPF-R2 and more complex kinematics.

\subsection{Wait Action Heatmaps}
\begin{figure}[tb]
    \centering
    \includegraphics[width=0.95\linewidth]{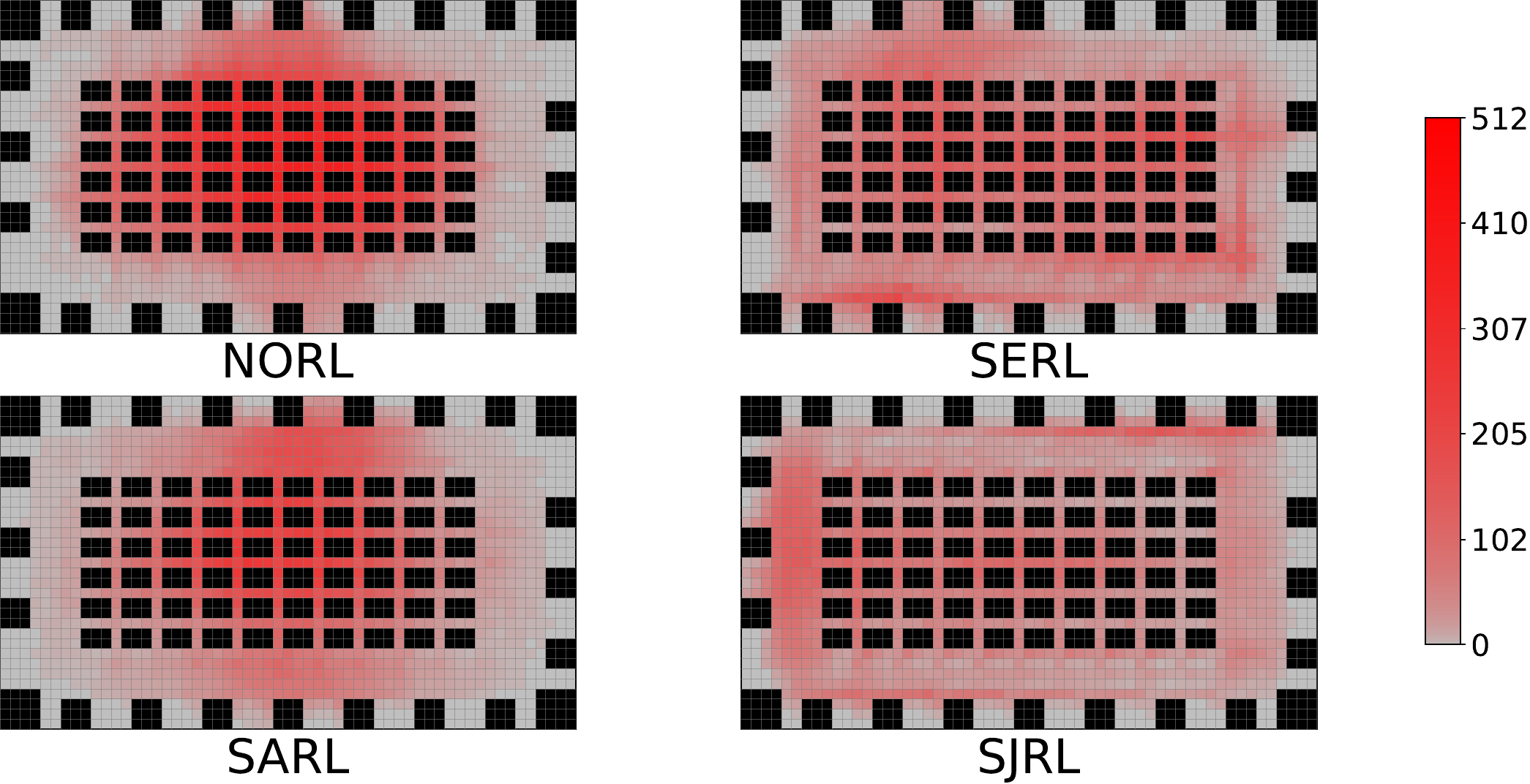}
    \caption{Heatmaps of average wait actions on the Warehouse map. Black cells denote obstacles; red intensity indicates wait frequency.}
    \label{fig:heatmap_warehouse}
    \vspace{-0.1cm}
\end{figure}

\begin{figure}[tb]
    \centering
    \includegraphics[width=0.95\linewidth]{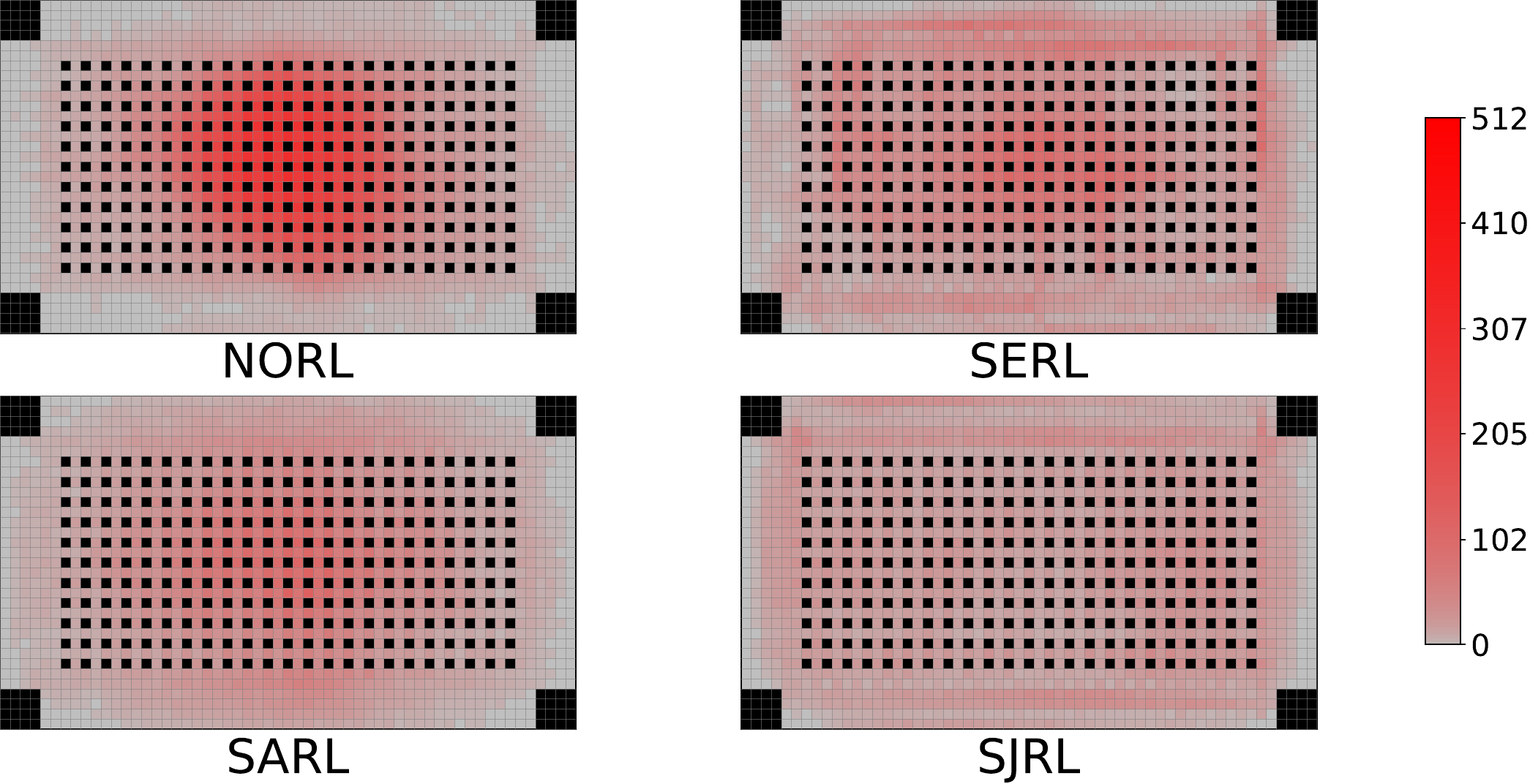}
    \caption{Heatmaps of average wait actions on the Sortation map. Black cells denote obstacles; red intensity indicates wait frequency.}
    \label{fig:heatmap_sortation}
    \vspace{-0.1cm}
\end{figure}

\begin{figure}[tb]
    \centering
    \includegraphics[width=0.95\linewidth]{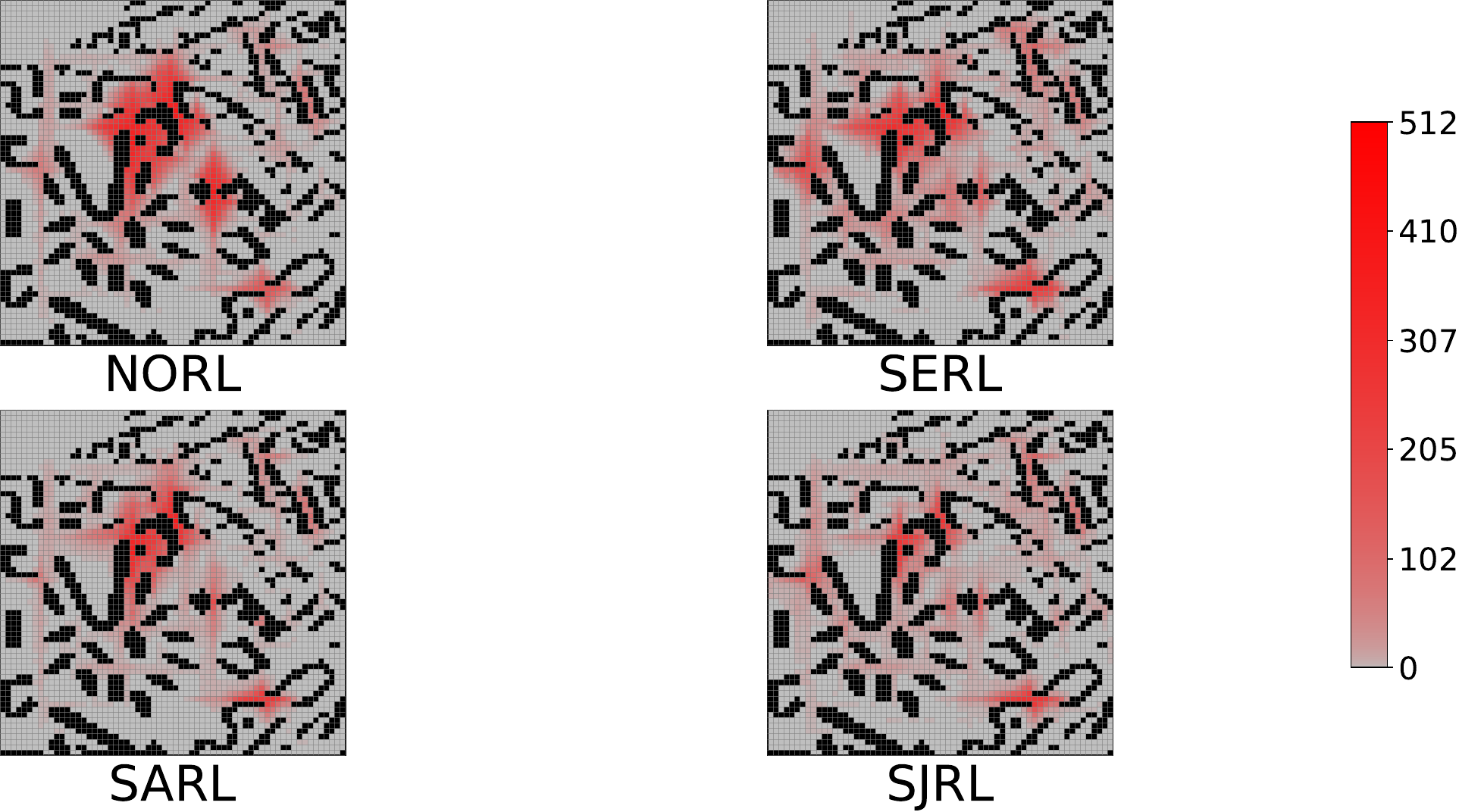}
    \caption{Heatmaps of average wait actions on the Paris map. Black cells denote obstacles; red intensity indicates wait frequency.}
    \label{fig:heatmap_paris}
    \vspace{-0.1cm}
\end{figure}

\begin{figure}[tb]
    \centering
    \includegraphics[width=0.95\linewidth]{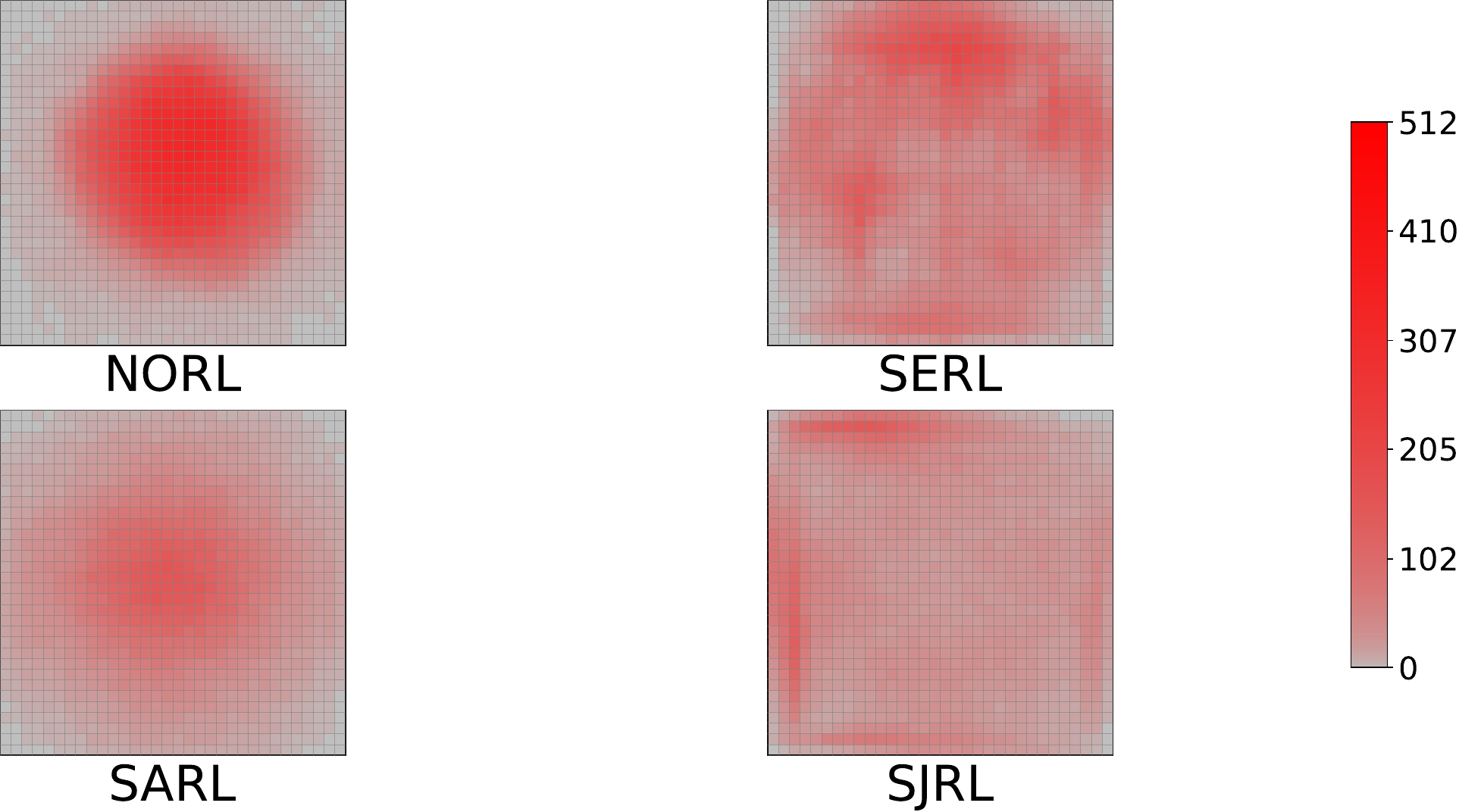}
    \caption{Heatmaps of average wait actions on the Empty map. Black cells denote obstacles; red intensity indicates wait frequency.}
    \label{fig:heatmap_emtpy}
    \vspace{-0.1cm}
\end{figure}

\begin{figure}[tb]
    \centering
    \includegraphics[width=0.95\linewidth]{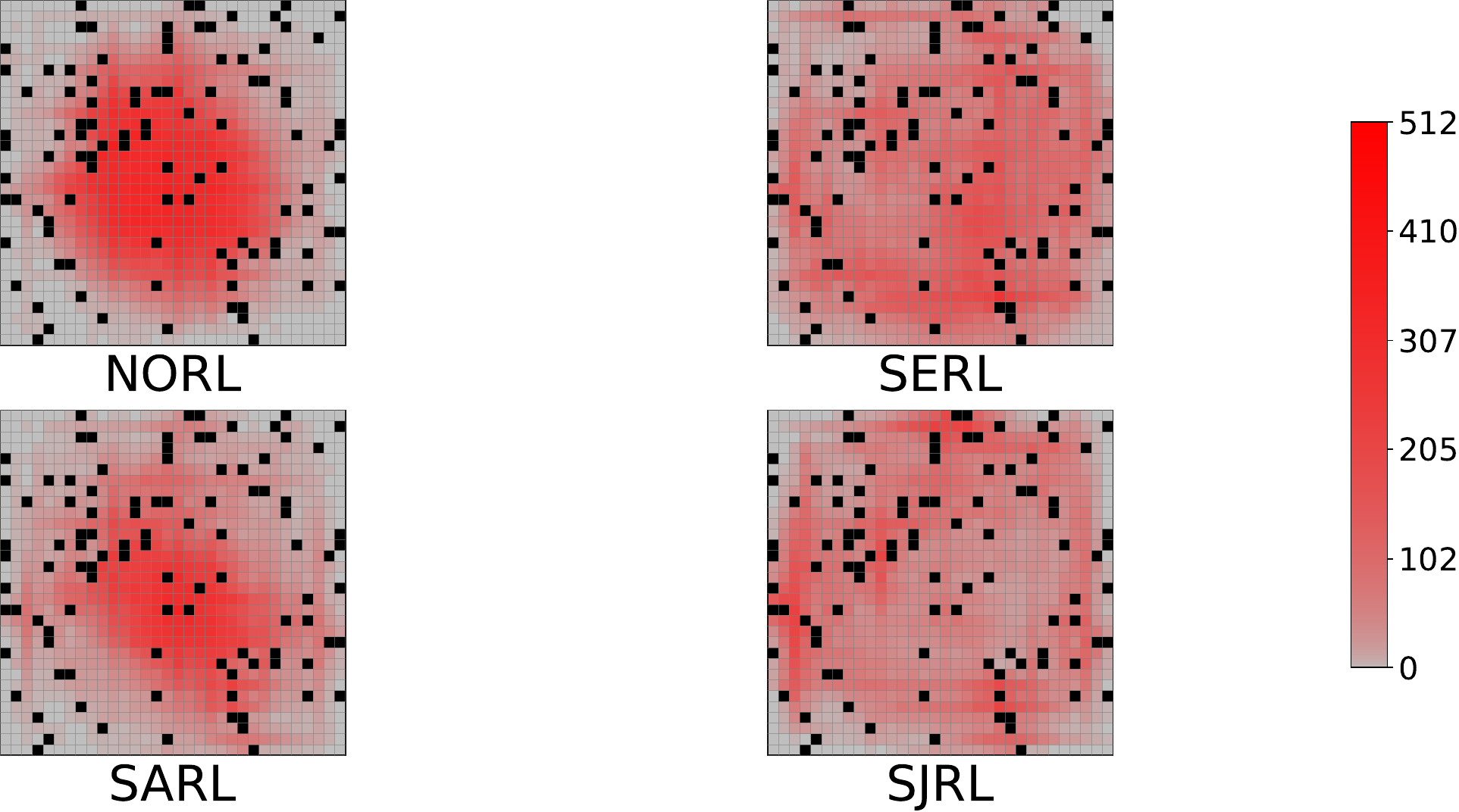}
    \caption{Heatmaps of average wait actions on the Random-10 map. Black cells denote obstacles; red intensity indicates wait frequency.}
    \label{fig:heatmap_random-10}
    \vspace{-0.1cm}
\end{figure}

\begin{figure}[tb]
    \centering
    \includegraphics[width=0.95\linewidth]{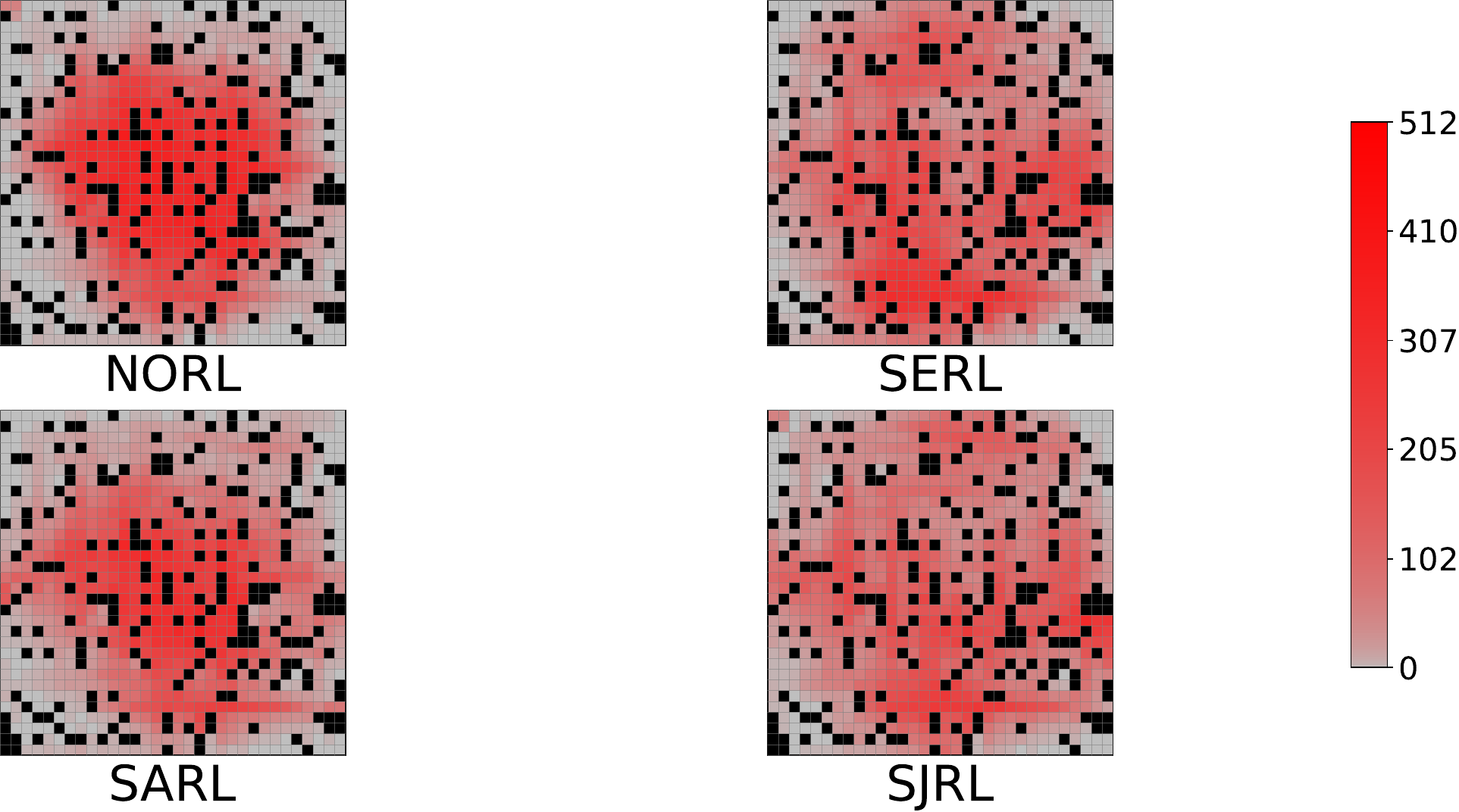}
    \caption{Heatmaps of average wait actions on the Random-20 map. Black cells denote obstacles; red intensity indicates wait frequency.}
    \label{fig:heatmap_random-20}
    \vspace{-0.1cm}
\end{figure}

In~\Cref{introduction}, we present the wait-action heatmaps only for the Sortation map in~\Cref{fig:heatmap}. Here, we provide the corresponding heatmaps for all maps in~\Cref{fig:heatmap_warehouse,fig:heatmap_sortation,fig:heatmap_paris,fig:heatmap_emtpy,fig:heatmap_random-10,fig:heatmap_random-20}. The same conclusions can be drawn from these visualizations: SARL alleviates congestion locally, SERL balances traffic globally, and SJRL combines the benefits of both.

\end{document}